\documentclass[sigconf,nonacm]{acmart}
\usepackage{caption}
\usepackage{multirow}
\usepackage{xcolor}
\usepackage{graphicx}
\usepackage{epstopdf}
\usepackage{url}
\usepackage{comment}
\usepackage{graphicx}
\usepackage{caption}
\usepackage{subcaption}
\usepackage{amsmath} 
\usepackage{tabularx}
\usepackage{ulem}
\usepackage{amssymb}
\AtBeginDocument{%
  \providecommand\BibTeX{{%
  \normalfont 
  \kern-0.5em{\scshape i\kern-0.25em b}\kern-0.8em\TeX}
  }}
\graphicspath{{./plots/GraphRAG Plots}}
\begin{document}

\title{HybridRAG: Integrating Knowledge Graphs and Vector Retrieval Augmented Generation for Efficient Information Extraction}

%%
%% The "author" command and its associated commands are used to define
%% the authors and their affiliations.
%% Of note is the shared affiliation of the first two authors, and the
%% "authornote" and "authornotemark" commands
%% used to denote shared contribution to the research.

%\author{Anonymous authors}
%\affiliation{%
%\institution{Anonymous}
%\city
%\country(Anonymous)
%}

% \author{Anonymous2}
% % \email{Anonymous@}
% \affiliation{%
% 	% \institution{Anonymous}
% 	% \city{New York}
% 	% \state{NY}
% 	\country{Anonymous}
% }
% \author{Anonymous3}
% % \email{Anonymous@}
% \affiliation{%
% 	% \institution{Anonymous}
% 	% \city{New York}
% 	% \state{NY}
% 	\country{Anonymous}
% }
% \author{Anonymous4}
% % \email{Anonymous@}
% \affiliation{%
% 	% \institution{Anonymous}
% 	% \city{New York}
% 	% \state{NY}
% 	\country{Anonymous}
% }
% \author{Anonymous5}
% % \email{Anonymous@}
% \affiliation{%
% 	% \institution{Anonymous}
% 	% \city{New York}
% 	% \state{NY}
% 	\country{Anonymous}
% }
% \author{Anonymous6}
% \affiliation{%
% 	% \institution{Anonymous}
% 	% \city{New York}
% 	% \state{NY}
% 	\country{Anonymous}
% }
% \renewcommand{\shortauthors}{Anonymous et al.}

%\begin{comment}
\author{Bhaskarjit Sarmah}
\email{bhaskarjit.sarmah@blackrock.com}
\affiliation{
  \institution{BlackRock, Inc.}
  \city{Gurugram}
  \country{India}
}
\author{Benika Hall}
\email{bhall@nvidia.com}
\affiliation{
  \institution{NVIDIA}
  \city{Santa Clara, CA}
  \country{USA}
}

\author{Rohan Rao}
\email{rohrao@nvidia.com}
\affiliation{
  \institution{NVIDIA}
  \city{Santa Clara, CA}
  \country{USA}
}

\author{Sunil Patel}
\email{supatel@nvidia.com}
\affiliation{
  \institution{NVIDIA}
  \city{Santa Clara, CA}
  \country{USA}
}

\author{Stefano Pasquali}
\email{stefano.pasquali@blackrock.com}
\affiliation{
  \institution{BlackRock, Inc.}
  \city{New York, NY}
  \country{USA}
}
\author{Dhagash Mehta}
\email{dhagash.mehta@blackrock.com}
\affiliation{
  \institution{BlackRock, Inc.}
  \city{New York, NY}
  \country{USA}
}
\renewcommand{\shortauthors}{Sarmah et al.}
%\end{comment}

\begin{abstract}
Extraction and interpretation of intricate information from unstructured text data arising in financial applications, such as earnings call transcripts, present substantial challenges to large language models (LLMs) even using the current best practices to use Retrieval Augmented Generation (RAG) (referred to as VectorRAG techniques which utilize vector databases for information retrieval) due to challenges such as domain specific terminology and complex formats of the documents. We introduce a novel approach based on a combination, called \textit{HybridRAG}, of the Knowledge Graphs (KGs) based RAG techniques (called GraphRAG) and VectorRAG techniques to enhance question-answer (Q\&A) systems for information extraction from financial documents that is shown to be capable of generating accurate and contextually relevant answers. Using experiments on a set of financial earning call transcripts documents which come in the form of Q\&A format, and hence provide a natural set of pairs of ground-truth Q\&As, we show that HybridRAG which retrieves context from both vector database and KG outperforms both traditional VectorRAG and GraphRAG individually when evaluated at both the retrieval and generation stages in terms of retrieval accuracy and answer generation. The proposed technique has applications beyond the financial domain.
\end{abstract}

\begin{comment}
\begin{CCSXML}
<ccs2012>
 <concept>
  <concept_id>10010520.10010553.10010562</concept_id>
  <concept_desc>Computer systems organization~Embedded systems</concept_desc>
  <concept_significance>500</concept_significance>
 </concept>
 <concept>
  <concept_id>10010520.10010575.10010755</concept_id>
  <concept_desc>Computer systems organization~Redundancy</concept_desc>
  <concept_significance>300</concept_significance>
 </concept>
 <concept>
  <concept_id>10010520.10010553.10010554</concept_id>
  <concept_desc>Computer systems organization~Robotics</concept_desc>
  <concept_significance>100</concept_significance>
 </concept>
 <concept>
  <concept_id>10003033.10003083.10003095</concept_id>
  <concept_desc>Networks~Network reliability</concept_desc>
  <concept_significance>100</concept_significance>
 </concept>
</ccs2012>
\end{CCSXML}

\ccsdesc[500]{Computer systems organization~Embedded systems}
\ccsdesc[300]{Computer systems organization~Redundancy}
\ccsdesc{Computer systems organization~Robotics}
\ccsdesc[100]{Networks~Network reliability}

%%
%% Keywords. The author(s) should pick words that accurately describe
%% the work being presented. Separate the keywords with commas.
\keywords{Company similarity, Natural Language Processing}

\end{comment}

\maketitle

\section{Introduction}
For the financial analyst, it is crucial to extract and analyze information from unstructured data sources like news articles, earnings reports, and other financial documents to have at least some chance to be on the better side of potential information asymmetry. These sources hold valuable insights that can impact investment decisions, market predictions, and overall sentiment. However, traditional data analysis methods struggle to effectively extract and utilize this information due to its unstructured nature. Large language models (LLMs) \cite{mikolov2013, vaswani2017attention, yang2020finbert, wu2023bloomberggpt} have emerged as powerful tools for financial services and investment management. Their ability to process and understand vast amounts of textual data makes them invaluable for tasks such as sentiment analysis, market trend predictions, and automated report generation. Specifically, extracting information from annual reports and other financial documents can greatly enhance the efficiency and accuracy of financial analysts \cite{nie2024survey}. A robust information extraction system can help analysts quickly gather relevant data, identify market trends, and make informed decisions, leading to better investment strategies and risk management \cite{zhao2024revolutionizingfinancellmsoverview}.

Although LLMs have substantial potential in financial applications, there are notable challenges in using pre-trained models to extract information from financial documents outside their training data while also reducing hallucination \cite{ling2023domain,sarmah2023towards}. Financial documents typically contain domain-specific language, multiple data formats, and unique contextual relationships that general purpose-trained LLMs do not handle well. In addition, extracting consistent and coherent information from multiple financial documents can be challenging due to variations in terminology, format, and context across different textual sources. The specialized terminology and complex data formats in financial documents make it difficult for models to extract meaningful insights, in turn, causing inaccurate predictions, overlooked insights, and unreliable analysis, which ultimately hinder the ability to make well-informed decisions.

Current approaches to mitigate these issues include various Retrieval-Augmented Generation (RAG) techniques \cite{lewis2020retrieval}, which aim to improve the performance of LLMs by incorporating relevant retrieval techniques. VectorRAG (the traditional RAG techniques that are based on vector databases) focuses on improving Natural Language Processing (NLP) tasks by retrieving relevant textual information to support the generation tasks. VectorRAG excels in situations where context from related textual documents is crucial for generating meaningful and coherent responses \cite{lewis2020retrieval, izacard2021leveraging, guu2020realm}. RAG-based methods ensure the LLMs generate relevant and coherent responses that are aligned with the original input query. However, for financial documents, these approaches have significant challenges as a standalone solution. For instance, traditional RAG systems often use paragraph-level chunking techniques, which assume the text in those documents are uniform in length. This approach neglects the hierarchical nature of financial statements and can result in the loss of critical contextual information for an accurate analysis\cite{jimeno2024financial, superacc2024rag}. Due to the complexities in analyzing financial documents, the quality of the LLM retrieved-context from a vast and heterogeneous corpus can be inconsistent, leading to inaccuracies and incomplete analyses. These challenges demonstrate the need for more sophisticated methods that can effectively integrate and process the detailed and domain-specific information found in financial documents, ensuring more reliable and accurate results for informed decision-making.

Knowledge graphs (KGs) \cite{ji2021survey} may provide a different point of view to looking at the financial documents where the documents are viewed as a collection of triplets of entities and their relationships as depicted in the text of the documents. KGs have become a pivotal technology in data management and analysis, providing a structured way to represent knowledge through entities and relationships and have been widely adopted in various domains, including search engines, recommendation systems, and biomedical research \cite{paulheim2017knowledge, hogan2021knowledge, ehrlinger2016towards}. The primary advantage of KGs lies in their ability to offer a structured representation, which facilitates efficient querying and reasoning. However, building and maintaining KGs and integrating data from different sources, such as documents, news articles, and other external sources, into a coherent knowledge graph poses significant challenges.

The financial services industry has recognized the potential of KGs in enhancing data integration of heterogeneous data sources, risk management, and predictive analytics \cite{li2024findkg,findkg2022,de2018financial, petrova2020knowledge, liu2019utilizing}. Financial KGs integrate various financial data sources such as market data, financial reports, and news articles, creating a comprehensive view of financial entities and their relationships. This unified view improves the accuracy and comprehensiveness of financial analysis, facilitates risk management by identifying hidden relationships, and supports advanced predictive analytics for more accurate market predictions and investment decisions. However, handling large volumes of financial data and continuously updating the knowledge graph to reflect the dynamic nature of financial markets can be challenging and resource-intensive.

GraphRAG (Graph-based Retrieval-Augmented Generation) \cite{edge2024local,yao2021retrieval, zhao2022graph, lin2020kgpt,gao2023retrieval,procko2024graph} is a novel approach that leverages knowledge graphs (KGs) to enhance the performance of NLP tasks such as Q\&A systems. By integrating KGs with RAG techniques, GraphRAG enables more accurate and context-aware generation of responses based on the structured information extracted from financial documents. But GraphRAG generally underperforms in abstractive Q\&A tasks or when there is not explicit entity mentioned in the question. 

In the present work, we propose a combination of VectorRAG and GraphRAG, called HybridRAG, to retrieve the relevant information from external documents for a given query to the LLM that brings advantages of both the RAGs together to provide demonstrably more accurate answers to the queries.

%Advantages:

%Improved Data Integration: Knowledge graphs facilitate the integration of diverse financial data sources, enhancing the overall analytical capabilities.
%Risk Management: By revealing hidden relationships and patterns, knowledge graphs can help in identifying and mitigating risks.
%Predictive Analytics: Supports advanced predictive analytics by structuring data in a way that makes it easier to analyze.

\subsection{Prior Work and Our Contribution}
VectorRAG has been extensively investigated in the recent years and focuses on enhancing NLP tasks by retrieving relevant textual information to support generation processes \cite{lewis2020retrieval, izacard2021leveraging, guu2020realm,gao2023retrieval}. However, the effectiveness of the retrieval mechanism across multiple documents and longer contexts poses a significant challenge in extracting relevant responses. GraphRAG combines the capabilities of KGs with RAG to improve traditional NLP tasks \cite{yao2021retrieval, zhao2022graph, lin2020kgpt}. Within our implementations of both VectorRAG GraphRAG techniques, we explicitly add information on the metadata of the documents that is also shown to improve the performance of VectorRAG \cite{sarmah2023towards}.

To the best of our knowledge the present work is the first work that proposes a RAG approach that is hybrid of both VectorRAG and GraphRAG and demonstrates its potential of more effective analysis and utilization of financial documents by leveraging the combined strengths of VectorRAG and GraphRAG. We also utilize a novel ground-truth Q\&A dataset extracted from publicly available financial call transcripts of the companies included in the Nifty-50 index which is an Indian stock market index that represents the weighted average of 50 of the largest Indian companies listed on the National Stock Exchange\footnote{\url{https://www.nseindia.com/products/content/equities/indices/nifty_50.htm}}.

\section{Methodology}
In this Section, we provide details of the proposed methodology by first discussing details of VectorRAG, then methodologies of constructing KGs from given documents and our proposed methodology of GraphRAG and finally the HybridGraph technique.

\subsection{VectorRAG}
The traditional RAG \cite{lewis2020retrieval} process begins with a query that is related to the information possessed within external document(s) that are not a part of the training dataset for the LLM. This query is used to search an external repository, such as a vector database or indexed corpus, to fetch relevant documents or passages containing useful information. These retrieved documents are then fed back into the LLM as additional context. Hence, in turn, for the given query, the language model generates a response based not only on its internal training data but also by incorporating the retrieved external information. This integration ensures that the generated content is grounded in more recent and verifiable data, improving the accuracy and contextual relevance of the responses. By combining the retrieval of external information with the generative capabilities of large language models, RAG enhances the overall quality and reliability of the generated text.

In traditional VectorRAG, the given external documents are divided into multiple chunks because of the limitation of context size of the language model. Those chunks are converted into embeddings using an embeddings model and then stored into a vector database. After that, the retrieval component performs a similarity search within the vector database to identify and rank the chunks most relevant to the query. The top-ranked chunks are retrieved and aggregated to provide context for the generative model.

Then, the generative model takes this retrieved context along with the original query and synthesizes a response. Thus, it merges the real-time information from the retrieved chunks with its pre-existing knowledge, ensuring that the response is both contextually relevant and detailed.

\begin{figure}[h]
\includegraphics[width=8cm]{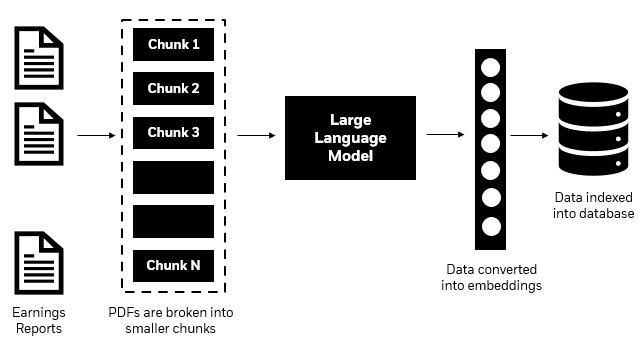}
\caption{A schematic diagram describing the vector database creation of a RAG application.}
\label{fig:chunking}
\end{figure}

The schematic diagram in Figure \ref{fig:chunking} provides details on the part of RAG that generates vector database from given external documents in the traditional VectorRAG methodology where we also include explicit reference of metadata \cite{sarmah2023towards}. Section \ref{sec:vectorRAG} will provide implementation details for our experiments.

\subsection{Knowledge Graph Construction}
A KG is a structured representation of real-world entities, their attributes, and their relations, usually stored in a graph database or a triplet store, i.e., a KG consists of nodes that represent entities and edges that represent relations, as well as labels and attributes for both. A graph triplet is a basic unit of information in a KG, consisting of a subject, a predicate, and an object.

In most methodologies to build a KG from given documents, three main steps are involved: knowledge extraction, knowledge improvement, and knowledge adaptation \cite{zhong2023comprehensive}. Here, we do not use knowledge adaptation and treat the KGs as static graphs.

\noindent\textbf{Knowledge Extraction:-} This step aims to extract structured information from unstructured or semi-structured data, such as text, databases, and existing ontologies. The main tasks in this step are entity recognition, relationship extraction, and co-reference resolution.
Entity recognition and relationship extraction techniques use typical NLP tasks to identify entities and their relationships from textual sources \cite{mondal2021end}. Coreference resolution identifies and connects different references of the same entity, keeping coherence within the knowledge graph. For example, if the text refers to a company as both "the company" and "it", coreference resolution can link these mentions to the same entity node in the graph.

\noindent\textbf{Knowledge Improvement:-} This step aims to enhance the quality and completeness of the KG by removing redundancies and addressing gaps in the extracted information. The primary tasks in this step are KG completion and fusion. KG completion technique infers missing entities and relationships within the graph using methods such as link prediction and entity resolution. Link prediction predicts the existence and type of a relation between two entities based on the graph structure and features, while entity resolution matches and merges different representations of the same entity from different sources.

Knowledge fusion combines information from multiple sources to create a coherent and unified KG. This involves resolving conflicts and redundancies among the sources, such as contradictory or duplicate facts, and aggregating or reconciling the information based on rules, probabilities, or semantic similarity.

\begin{comment}
\noindent\textbf{Knowledge Adaptation:-} This step aims to update and maintain the KG as new information becomes available. The main tasks in this step are to create dynamic, temporal and event KGs \cite{zhong2023comprehensive}. Dynamic KGs adapt to new information and changes over time, ensuring its currency and relevance. This involves detecting and incorporating new entities, relationships, and attributes, as well as deleting or modifying outdated or incorrect ones, whereas Temporal and Event KG techniques capture the temporal aspects and event-based information of the KG, enabling more sophisticated reasoning over time. This involves adding timestamps and sequences to the entities and edges or relationships, as well as modeling events as complex entities with various attributes.
\textcolor{red}{\sout{@Bhaskarjit, are we using Knowledge adaptation step in HybridRAG?}}
\textcolor{blue}{No we are not using this step here. Our KG is static and not Dynamic}
\end{comment}

We utilized a robust methodology for creating KG triplets from unstructured text data, specifically focusing on corporate documents such as earnings call transcripts, adapted from Ref.~\cite{li2023findkg,li2024findkg}. This process involves several interconnected stages, each designed to extract, refine, and structure information effectively.

We implement a two-tiered LLM chain for content refinement and information extraction. The first tier employs an LLM to generate an abstract representation of each document chunk. This refinement process is crucial as it distills the essential information while preserving the original meaning and key relationships between concepts that serves as a more focused input for subsequent processing, enhancing the overall efficiency and accuracy of our triplet extraction pipeline. The second tier of our LLM chain is dedicated to entity extraction and relationship identification. 

Both the steps are executed using carefully performed prompt engineering on a pre-trained LLM. A detailed discussion on implementation of the methodology will be provided in Section \ref{sec:Knowledge_Graph}
 
\subsection{GraphRAG}
KG based RAG \cite{edge2024local}, or GraphRAG, also begins with a query based on the user's input same as VectorRAG. The main difference between VectorRAG and GraphRAG lies in the retrieval part. The query here is now used to search the KG to retrieve relevant nodes (entities) and edges (relationships) related to the query. A subgraph, which consists of these relevant nodes and edges, is extracted from the full KG to provide context. This subgraph is then integrated with the language model's internal knowledge, by encoding the graph structure into embeddings that the model can interpret. The language model uses this combined context to generate responses that are informed by both the structured information from the KG and its pre-trained knowledge. Crucially, when responding to user queries about a particular company, we leveraged the metadata information to selectively filter and retrieve only those document segments pertinent to the queried company \cite{sarmah2023towards}. This integration helps ensure that the generated outputs are accurate, contextually relevant, and grounded in verifiable information.

\begin{figure}[h]
\includegraphics[width=8cm]{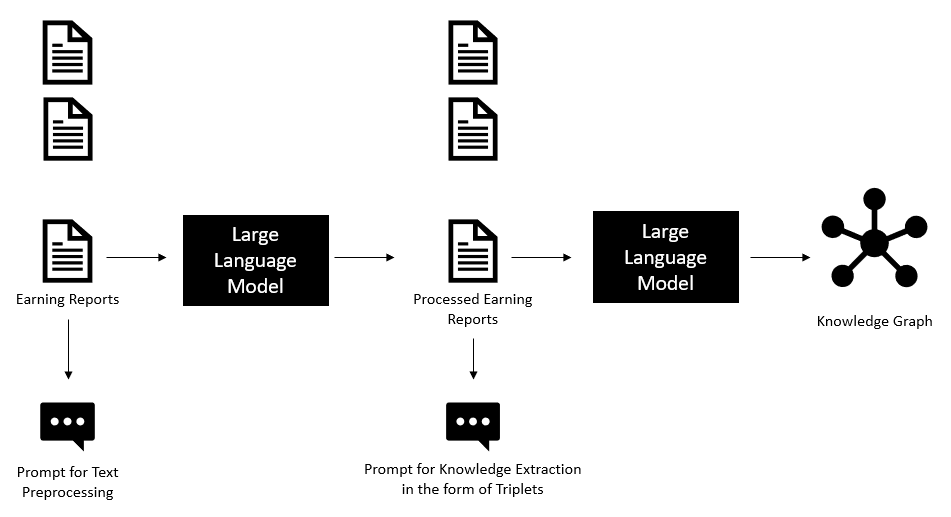}
\caption{A schematic diagram describing knowledge graph creation process of GraphRAG.}
\label{fig:graph_chunking}
\end{figure}

A schematic diagram of the retrieval methodology of GraphRAG is given in Figure \ref{fig:graph_chunking}. Here we first write a prompt to clean the data and then write another prompt in the second stage to create knowledge triplets along with metadata. It will be covered in more detail in section \ref{sec:Knowledge_Graph}

\subsubsection{HybridRAG}
For the HybridRAG technique, we propose to integrate the aforementioned two distinct RAG techniques: VectorRAG and GraphRAG. This integration involves a systematic combination of contextual information retrieved from both the traditional vector-based retrieval mechanism and the KG-based retrieval system, the latter of which was constructed specifically for this study.

The amalgamation of these two contexts allows us to leverage the strengths of both approaches. The VectorRAG component provides a broad, similarity-based retrieval of relevant information, while the GraphRAG element contributes structured, relationship-rich contextual data. This combined context is then utilized as input for a LLM to generate the final responses. Details on the implementation of the HybridRAG will be provided in Section \ref{sec:HybridRAG}.

\subsection{Evaluation Metrics}
To assess the efficacy of this integrated approach, we conducted a comparative analysis among the three approaches in a controlled experimental set up: VectorRAG, GraphRAG and HybridRAG. The responses generated using the combined VectorRAG and GraphRAG contexts were juxtaposed against those produced individually by VectorRAG and GraphRAG. This comparative evaluation aimed to discern potential improvements in response quality, accuracy, and comprehensiveness that might arise from the synergistic integration of these two RAG methodologies.

To objectively evaluate different RAG approaches (VectorRAG and GraphRAG in their case), Ref.~\cite{edge2024local} utilized metrics such as comprehensiveness (i.e., the amount of details the answer provides to cover all aspects and
details of the question?); diversity (i.e., the richness of the answer in providing different perspectives and insights
on the question); empowerment (i.e., the helpfulness of the answer to the reader understand and make informed judgements about the topic); and, directness (i.e., clearness of the answer in addressing the question). here, the LLM was provided tuples of question, target metric, and a pair of answers, and was asked to assess which answer was better according to the metric and why.

These metrics though compare the final generated answers, do not necessarily directly evaluate the retrieval and generation parts separately. Instead, here we implement a comprehensive set of evaluation metrics which are designed to capture different aspects of a given RAG system's output quality, focusing on faithfulness, answer relevance, and context relevance \cite{es2023ragas}. Each metric provides unique insights into the system's capabilities and limitations.

\subsubsection{Faithfulness} Faithfulness is a crucial metric that measures the extent to which the generated answer can be inferred from the provided context. Our implementation of the faithfulness metric involves a two-step process:

\noindent\textbf{Statement Extraction:-} We use an LLM to decompose the generated answer into a set of concise statements. This step is crucial for breaking down complex answers into more manageable and verifiable units. The prompt used for this step is:

"Given a question and answer, create one or more statements from each sentence in the given answer.
question: [question]
answer: [answer]".

\noindent\textbf{Statement Verification:-} For each extracted statement, we employ the LLM to determine if it can be inferred from the given context. This verification process uses the following prompt:

"Consider the given context and following statements, then determine whether they are supported by the information present in the context. Provide a brief explanation for each statement before arriving at the verdict (Yes/No). Provide a final verdict for each statement in order at the end in the given format. Do not deviate from the specified format.
statement: [statement 1]
...
statement: [statement n]".

The faithfulness score ($F$) is $F = |V| / |S|$, where $|V|$ is the number of supported statements and $|S|$ is the total number of statements.

\subsubsection{Answer Relevance:-} The answer relevance metric assesses how well the generated answer addresses the original question, irrespective of factual accuracy. This metric helps identify cases of incomplete answers or responses containing irrelevant information. Our implementation involves the following steps:

Question Generation: We prompt the LLM to generate n potential questions based on the given answer:

"Generate a question for the given answer.
answer: [answer]".

Then, we obtain embeddings for all generated questions and the original question using OpenAI's text-embedding-ada-002 model\footnote{\url{https://platform.openai.com/docs/guides/embeddings/embedding-models}}.
We then calculate the cosine similarity between each generated question's embedding and the original question's embedding.

Finally, the answer relevance score (AR) is computed as the average similarity across all generated questions: $AR = \frac{1}{n} \sum(sim(q, q_i))$, where $sim(q, q_i)$ is the cosine similarity between the embedding of the original question $q$ and the embeddings of each of the $n$ generated questions $q_i$.

\subsubsection{Context Precision}
It is a metric used to evaluate the relevance of retrieved context chunks in relation to a specified ground truth for a given question\footnote{\url{https://docs.ragas.io/en/stable/concepts/metrics/context_precision.html}}. It calculates the proportion of relevant items that appear in the top ranks of the context. The formula for context precision at K is the sum of the products of precision at each rank k and a binary relevance indicator v\_k, divided by the total number of relevant items in the top K results. Precision at each rank k is determined by the ratio of true positives at k to the sum of true positives and false positives at k. This metric helps in assessing how well the context supports the ground truth, aiming for higher scores which indicate better precision.

\subsubsection{Context Recall} 
It is a metric used to evaluate how well the retrieved context aligns with the ground truth answer, which is considered the definitive correct response\footnote{\url{https://docs.ragas.io/en/stable/concepts/metrics/context_recall.html}}. It is quantified by comparing each sentence in the ground truth answer to see if it can be traced back to the retrieved context. The formula for context recall is the ratio of the number of ground truth sentences that can be attributed to the context to the total number of sentences in the ground truth. Higher values, ranging from 0 to 1, indicate better alignment and thus better context recall. This metric is crucial for assessing the effectiveness of information retrieval systems in providing relevant context.

\section{Data Description}
Although there do exist some public financial datasets, none of them were suitable for the present experiments: e.g., FinQA \cite{chen2021finqa}, TAT-QA \cite{maia201818}, FIQA \cite{zhu2021tat}, FinanceBench \cite{islam2023financebench}, etc. datasets are limited to specific usecases such as benchmarking LLMs' abilities to perform complex numerical reasoning or sentiment analysis. On the other hand, FinTextQA \cite{chen2024fintextqa} dataset was not publicly available at the time of writing the present work. In addition, in most of these datasets, access to the actual documents from which the ground-truth Q\&As were created is not available, making it impossible to use them for our RAG techniques evaluation purposes. Hence, we resorted to a dataset of our own though through publicly available documents but such that we finally have access to both the actual financial documents and the ground-truth Q\&As. Datasets like FinanceBench\footnote{\url{https://huggingface.co/datasets/PatronusAI/financebench-test}} provides question-context-answer triplets but they are not useful  as here we are comparing VectorRAG, GraphRAG and HybridRAG and they do not provide the context generated from a KG. A recent paper \cite{edge2024local} has not made the KG and triplets constructed by their algorithm public to the best of our knowledge either. 

In short, there is no publicly available benchmark dataset to compare VectorRAG and GraphRAG either for financial or general domains to the best of our knowledge. Hence, we had to rely on our own dataset constructed as explained below.

We used transcripts from earnings calls of Nifty 50 constituents for
our analysis. The NIFTY 50 is popular index in the Indian stock market that represents the weighted average of 50 of the largest Indian companies listed on the National Stock Exchange (NSE). The dataset of the earning call documents of Nifty 50 companies is widely recognized in the investment realm and is esteemed as an authoritative and extensive collection
of earnings call transcripts. In our investigation, we focus on data
spanning the quarter ending in June, 2023 i.e. the earnings reports
for Q1 of the financial year 2024 (A financial year in India starts on the 1st April and ends in 31st March, so the quarter from 1st April to 30th June is the first quarter of 2024 for the Indian market). 

Our dataset encompasses 50
transcripts for this quarter, spanning over 50 companies within
Nifty 50 universe from diverse range of sectors including Infrastructure,
Healthcare, Consumer Durables, Banking, Automobile, Financial
Services, Energy - Oil \& Gas, Telecommunication, Consumer Goods,
Pharmaceuticals, Energy - Coal, Materials, Information Technology,
Construction, Diversified, Metals, Energy - Power and Chemicals
providing a substantial and diverse foundation for our study.

We start the data collection process focused on acquiring earnings reports from company websites within the Nifty 50 universe by developing and deploying a custom web scraping tool to navigate through the websites of each company within the Nifty 50 index, systematically retrieving the pertinent earnings reports for Q1 of the financial year 2024. By utilizing this web scraping approach, we aimed to compile a comprehensive dataset encompassing the earnings reports of the constituent companies.

Table \ref{tab:summary} summarizes basic statistics of the documents we will be experimenting with in the remainder of this work.

\begin{table}[ht]
\centering
\begin{tabular}{|l|c|}
\hline
Number of companies/documents & 50 \\
Average number of pages & 27 \\
Average number of questions & 16 \\
Average number of tokens & 60,000 \\
\hline
\end{tabular}
\caption{Summary Statistics for the call transcript documents used in the present work.}
\label{tab:summary}
\end{table}
\vspace{-0.25in}

These call transcripts documents consist of questions and answers between financial analysts and the company representatives for the respective companies, hence, there already exist certain Q\&A pairs within these documents along with additional text. We examined the earnings reports within the Nifty50 universe, systematically curated a comprehensive array of randomly selected 400 questions posed during the earnings calls from all the documents, and gathered the exact responses corresponding to these questions. These questions constitute the specific queries articulated by financial analysts to the management during these calls.

\section{Implementation Details}
In this Section, we provide details of implementation of the proposed methodology.

\subsection{Knowledge Graph Construction} \label{sec:Knowledge_Graph}
The initial phase of our approach centers on document preprocessing. We utilize the PyPDFLoader\footnote{\url{https://python.langchain.com/v0.1/docs/modules/data_connection/document_loaders/pdf/}} to import PDF documents, which are subsequently segmented into manageable chunks using the RecursiveCharacterTextSplitter. This chunking strategy employs a size of 2024 characters with an overlap of 204 characters, ensuring comprehensive coverage while maintaining context across segment boundaries.

Following the preprocessing stage, we implement the two-tiered language model chain for content refinement and information extraction. It is not possible to include the exact prompt here due to the limited space, but a baseline prompt can be found in Ref.~\cite{li2024findkg}.

\begin{table}[h]
\centering
\resizebox{\columnwidth}{!}{%
\begin{tabular}{|p{0.35\columnwidth}|p{0.55\columnwidth}|}
\hline
\textbf{Entity Type} & \textbf{Examples} \\
\hline
Companies and Corporations & Official names, abbreviations, informal references \\
\hline
Financial Metrics and Indicators & Revenue, profit margins, EBITDA \\
\hline
Corporate Executives and Key Personnel & CEOs, CFOs, board members \\
\hline
Products and Services & Tangible products and intangible services \\
\hline
Geographical Locations & Headquarters, operational regions, markets \\
\hline
Corporate Events & Mergers, acquisitions, product launches, earnings calls \\
\hline
Legal and Regulatory Information & Legal cases, regulatory compliance \\
\hline
\end{tabular}%
}
\caption{Entities extracted from earnings call transcripts}
\label{tab:corporate_info}
\end{table}

Table \ref{tab:corporate_info} summarizes details on entities extracted from the earning calls transcripts using our prompt based method. Concurrently, LLM identifies relationships between these entities using a curated set of verbs, capturing the nuanced interactions within the corporate narrative. A key improvement in our methodology is the enhanced prompt engineering to generate the structured output format for knowledge triplets. Each triplet is represented as a nested list ['h', 'type', 'r', 'o', 'type', 'metadata'], where 'h' and 'o' denote the head and object entities respectively, 'type' specifies the entity category, 'r' represents the relationship, and 'metadata' encapsulates additional contextual information. This format allows for a rich, multidimensional representation of information, facilitating more nuanced downstream analysis.

Our process incorporates several advanced features to enhance the quality and utility of the extracted triplets. Entity disambiguation techniques are employed to consolidate different references to the same entity, improving consistency across the KG. We also prioritize conciseness in entity representation, aiming for descriptions of less than four words where possible, which aids in maintaining a clean and navigable graph structure.

The extraction pipeline is applied iteratively to each document chunk, with results aggregated to form a comprehensive knowledge graph representation of the entire document, allowing for scalable processing of large documents while maintaining local context within each chunk. We have added explicit instruction on obtaining metadata following Ref.~\cite{sarmah2023towards} for both VectorRAG and GraphRAG.

Finally, we implement a data persistence strategy, converting the extracted triplets from their initial string format to Python data structures and storing them in a pickle file. This facilitates easy retrieval and further manipulation of the knowledge graph data in subsequent analysis stages.

%Figure \ref{fig:KG_financial_services} shows an example KG for entities and relationships extracted using our methodology for all the Nifty 50 companies that are in the Financial Services sector.

Our methodology represents a significant advancement in automated knowledge extraction from corporate documents. By combining advanced NLP techniques with a structured approach to information representation, we create a rich, queryable knowledge base that captures the complex relationships and key information present in corporate narratives. This approach opens up new possibilities for financial analysis and automated reasoning in the business domain that will be explored further in the future.

\begin{comment}
\begin{figure}[h]
\includegraphics[width=6cm, height=6cm]{plots/GraphRAG Plots/financial_services.jpg}
\caption{The Knowledge Graph created using the our methodology from earning call documents of all  companies from the Financial Services sector companies from the Nifty 50 universe.}
\label{fig:KG_financial_services}
\end{figure}

\begin{figure}[h]
\includegraphics[width=6cm, height=6cm]{plots/GraphRAG Plots/information_technology.jpg}
\caption{Information Technology}
\label{fig:KG_information_tech}
\end{figure}
\end{comment}

\subsection{VectorRAG}\label{sec:vectorRAG}
Our methodology builds upon the concept of RAG \cite{lewis2020retrieval} which allows for the creation of a system that can provide context-aware, accurate responses to queries about company financial information, leveraging both the power of large language models and the efficiency of semantic search.

At the core of our system is a Pinecone vector database\footnote{\url{https://www.pinecone.io/}}, which serves as the foundation for our information retrieval process. We employ OpenAI's text-embedding-ada-002 model to transform textual data from earnings call transcripts into high-dimensional vector representations. This vectorization process enables semantic similarity searches, significantly enhancing the relevance and accuracy of retrieved information. Table \ref{table:config} provides summary of the configuration of the set up in use for our experiments.

\begin{table}[h]
\centering
\begin{tabular}{|l|l|}
\hline
\textbf{LLM}                         & GPT-3.5-Turbo              \\ \hline
\textbf{LLM Temperature}   & 0 \\ \hline
\textbf{Embedding Model}             & text-embedding-ada-002     \\ \hline
\textbf{Framework}                   & LangChain                  \\ \hline
\textbf{Vector Database}             & Pinecone                   \\ \hline
\textbf{Chunk Size}                  & 1024                       \\ \hline
\textbf{Chunk Overlap}               & 0                          \\ \hline
\textbf{Maximum Output Tokens}       & 1024                       \\ \hline
\textbf{Chunks for Similarity Algorithm} & 20                        \\ \hline
\textbf{Number of Context Retrieved} & 4                          \\ \hline
\end{tabular}
\caption{VectorRAG Configuration}
\label{table:config}
\end{table}

The Q\&A pipeline is constructed using the LangChain framework\footnote{\url{https://docs.smith.langchain.com/old/cookbook/hub-examples/retrieval-qa-chain}}. The begins with a context retrieval step, where we query the Pinecone vector store to obtain the most relevant document chunks for a given question. This retrieval process is fine-tuned with specific filters for quarters, years, and company names, ensuring that the retrieved information is both relevant and current.

Following retrieval, we implement a context formatting step that consolidates the retrieved document chunks into a coherent context string. This formatted context serves as the informational basis for the language model's response generation. We have developed a sophisticated prompt template, that instructs the language model to function as an expert Q\&A system, emphasizing the importance of utilizing only the provided context information and avoiding direct references to the context in the generated responses.

For the core language processing task, we integrate OpenAI's GPT-3.5-turbo model which processes the formatted context and query to generate natural language responses that are informative, coherent, and contextually appropriate.

To evaluate the performance of our system, we developed a comprehensive framework that includes the preparation of a custom dataset of question-answer pairs specific to each company's earnings call. Our system processes each question in this dataset, generating answers based on the retrieved context. The evaluation results, including the original question, generated answer, retrieved contexts, and ground truth, are compiled into structured formats (CSV and JSON) to facilitate further analysis. The outputs generated by our system are stored in both CSV and JSON formats, enabling easy integration with various analysis tools and dashboards. This approach facilitates both quantitative performance metrics and qualitative assessment of the system's responses, providing a comprehensive view of its effectiveness.

By parameterizing company names, quarters, and years, we can easily adapt the system to different datasets and time periods. This design choice allows for seamless integration of new data and expansion to cover multiple companies and earnings calls.

\subsection{GraphRAG}\label{sec:GraphRAG}
For GraphRAG, we developed an Q\&A system specifically designed for corporate earnings call transcripts. Our implementation of GraphRAG leverages several key components and techniques:

\begin{table}[h]
\centering
\begin{tabular}{|l|l|}
\hline
\textbf{LLM}               & GPT-3.5-Turbo \\ \hline
\textbf{LLM Temperature}   & 0 \\ \hline
\textbf{Framework}         & LangChain     \\ \hline
\textbf{KG Manipulation}   & Networkx      \\ \hline
\textbf{Chunk Size}   & 1024      \\ \hline
\textbf{Chunk Overlap}   & 0      \\ \hline
\textbf{Number of Triplets}   & 13950      \\ \hline
\textbf{Number of nodes}   & 11405         \\ \hline
\textbf{Number of edges}   & 13883         \\ \hline
\textbf{DFS Depth}   & 1         \\ \hline
\end{tabular}
\caption{GraphRAG Configuration}
\label{table:kg_config}
\end{table}

\noindent\textbf{Knowledge Graph Construction:-}
We begin by constructing a KG from a set of knowledge triplets extracted from corporate documents using the prompt engineering based methodology as described in Section \ref{sec:Knowledge_Graph}. These triplets, stored in a pickle file, represent structured information in the form of subject-predicate-object relationships. We use the NetworkxEntityGraph class from the LangChain library to create and manage this graph structure. Each triple is added to the graph, which encapsulates the head entity, relation, tail entity, and additional metadata.

We implement the Q\&A functionality using the GraphQAChain class from LangChain. This chain combines the KG with an LLM (in our case, OpenAI's GPT-3.5-turbo) to generate responses. The GraphQAChain traverses the KG to find relevant information and uses the language model to formulate coherent answers based on the retrieved context. A summary of configuration of our LLM models and other libraries used for GraphRAG is shown in Table \ref{table:kg_config}.

In KG as the information is stored in the form of entities and relationships and there can be multiple relations emanating from one single entity, in this experiment, to extract relevant information from the KG, we employ a depth-first search strategy constrained by a depth of one from the specified entity.

To prepare for assessing the performance of our GraphRAG system, we follow the below steps:
\noindent\textbf{Dataset Preparation:-} We use a pre-generated CSV file containing question-answer pairs specific to the earnings call transcript.

\noindent\textbf{Iterative Processing:-} For each question in the dataset, we run the GraphQAChain to generate an answer.

\noindent\textbf{Result Compilation:-} We compile the results, including the original questions, generated answers, retrieved contexts, and ground truth answers, into a structured format.

Finally, the evaluation results are saved in both CSV and JSON formats for further analysis and comparison. We then fed these outputs into our RAG evaluation pipeline. For each Q\&A pair in our dataset, we compute all three metrics: faithfulness, answer relevance, context precision and context recall.

\subsection{HybridRAG}\label{sec:HybridRAG}
For the proposed HybridRAG technique, upon obtaining all the contextual information from VectorRAG and GraphRAG, we concatenate these contexts to form a unified context utilizing both techniques. This combined context is then fed into the answer generator model to produce a response. The context used for response generation is relatively larger, which affects the precision of the generated response. The context from VectorRAG is appended first, followed by the context from GraphRAG. Consequently, the precision of the generated answer depends on the source context. If the answer is generated from the GraphRAG context, it will have lower precision, as the GraphRAG context is added last in the sequence of contexts provided to the answer generator model, and vice versa.

\section{Results}
We evaluate both the retrieval and generation parts of RAG for the three different RAG pipelines. Evaluating the RAG outputs is also an active area of research there is no standard tool which is universally accepted as of yet, though we use a currently popular framework RAGAS\cite{es2023ragas} to evaluate the three RAG pipelines in the present work where we have modified them a bit to make more precise comparisons.

The results of our comparative analysis reveal notable differences in performance among VectorRAG, GraphRAG, and HybridRAG approaches as summarized in Table \ref{table:metrics}. In terms of Faithfulness, GraphRAG and HybridRAG demonstrated superior performance, both achieving a score of 0.96, while VectorRAG trailed slightly with a score of 0.94. Answer relevancy scores varied across the methods, with HybridRAG outperforming the others at 0.96, followed by VectorRAG at 0.91, and GraphRAG at 0.89. Context precision was highest for GraphRAG at 0.96, significantly surpassing VectorRAG (0.84) and HybridRAG (0.79). However, in context recall, both VectorRAG and HybridRAG achieved perfect scores of 1, while GraphRAG lagged behind at 0.85. 

Overall, these results suggest that GraphRAG offers improvements over VectorRAG, particularly in faithfulness and context precision. Furthermore, HybridRAG emerges as the most balanced and effective approach, outperforming both VectorRAG and GraphRAG in key metrics such as faithfulness and answer relevancy, while maintaining high context recall. 

The relatively lower context precision observed for HybridRAG (0.79) can be attributed to its unique approach of combining contexts from both VectorRAG and GraphRAG methods. While this integration allows for more comprehensive information retrieval, it also introduces additional content that may not align precisely with the ground truth, thus affecting the context precision metric. Despite this trade-off, HybridRAG's superior performance in faithfulness, answer relevancy, and context recall underscores its effectiveness. When considering the overall evaluation metrics, HybridRAG emerges as the most promising approach, balancing high-quality answers with comprehensive context retrieval.

Overall GraphRAG performs better in extractive questions compared to VectorRAG. And VectorRAG does better in abstractive questions where information is not explicitly mentioned in the raw data. And also GraphRAG sometimes fails to answer questions correctly whenever there is no entity explicitly mentioned in the question. So HybridRAG does a good job overall, as whenever VectorRAG fails to fetch correct context in extractive questions it falls back to GraphRAG to generate the answer. And whenever GraphRAG fails to fetch correct context in abstractive questions it falls back to VectorRAG to generate the answer.

\begin{table}[h]
\centering
\begin{tabular}{|l|c|c|c|}
\hline
                       & \textbf{VectorRAG} & \textbf{GraphRAG} & \textbf{HybridRAG} \\ \hline
F   & 0.94          & 0.96              & 0.96               \\ \hline
AR & 0.91          & 0.89              & 0.96               \\ \hline
CP & 0.84          & 0.96              & 0.79               \\ \hline
CR    & 1             & 0.85              & 1                  \\ \hline
\end{tabular}
\caption{Performance Metrics for Different RAG Pipelines. Here, F, AR, CP and CR refer to Faithfulness, Answer Relevence, Context Precision and Context Recall.}
\label{table:metrics}
\end{table}

\begin{comment}
\begin{figure}[h]
\includegraphics[width=8cm]{plots/GraphRAG Plots/graph_rag.png}
\caption{Performance Metrics for Different RAG Pipelines}
\label{fig:chunking}
\end{figure}
\end{comment}

\section{Conclusion and Future Directions}
Among the current approaches to mitigate issues regarding information extraction from external documents that were not part of training data for the LLM, Retrieval Augmented Generation (RAG) techniques have emerged as the most popular ones that aim to improve the performance of LLMs by incorporating relevant retrieval mechanisms. RAG methods enhance the LLMs' capabilities by retrieving pertinent documents or text to provide additional context during the generation process. However, these approaches encounter significant limitations when applied to the specialized and intricate domain of financial documents. Furthermore, the quality of the retrieved context from a vast and heterogeneous corpus can be inconsistent, leading to inaccuracies and incomplete analyses. These challenges highlight the need for more sophisticated methods that can effectively integrate and process the detailed and domain-specific information found in financial documents, ensuring more reliable and accurate outputs for informed decision-making.

In the present work, we have introduced a novel approach that significantly advances the field of information extraction from financial documents through the development of a hybrid RAG system. This system, called HybridRAG, which integrates the strengths of both Knowledge Graphs (KGs) and advanced language models, represents a leap forward in our ability to extract and interpret complex information from unstructured financial texts. The hybrid RAG system, by combining traditional vector-based RAG and KG-based RAG, has shown superior performance in terms of retrieval accuracy and answer generation, marking a pivotal step towards more effective financial analysis tools.

Through a comparative analysis using objecive evaluation metrics, we have highlighted the distinct performance advantages of the HybridRAG approach over its vector-based and KG-based counterparts. The HybridRAG system excels in faithfulness, answer relevancy, and context recall, demonstrating the benefits of integrating contexts from both VectorRAG and GraphRAG methods, despite potential trade-offs in context precision.

The implications of this research extend beyond the immediate realm of financial analysis. By developing a system capable of understanding and responding to nuanced queries about complex financial documents, we pave the way for more sophisticated AI-assisted financial decision-making tools that could potentially democratize access to financial insights, allowing a broader range of stakeholders to engage with and understand financial information.

Future directions for this research include expanding the system to handle multi-modal inputs, incorporating numerical data analysis capabilities, and developing more sophisticated evaluation metrics that capture the nuances of financial language and the accuracy of numerical information in the responses. Additionally, exploring the integration of this system with real-time financial data streams could further enhance its utility in dynamic financial environments.

%\begin{comment}
\section{Acknowledgement}
The views expressed here are those of the authors alone and not of BlackRock, Inc or NVIDIA. We are grateful to Emma Lind for her invaluable support for this collaboration.
%BlackRock, Inc.
%\bibliographystyle{unsrt}
%\end{comment}

%\bibliographystyle{ACM-Reference-Format}
\bibliography{sample-base}
\end{document}